# Linking Artificial Intelligence Principles

Yi Zeng[1,2,3], Enmeng Lu[1,*], Cunqing Huangfu[1,*]

[1] Institute of Automation, Chinese Academy of Sciences
[2] School of Artificial Intelligence, University of Chinese Academy of Sciences
[3] Berggruen Institute China Center, Peking University
{yi.zeng, enmeng.lu, cunqing.huangfu}@ia.ac.cn

**Abstract**

Artificial Intelligence principles define social and ethical considerations to develop future AI. They come from research institutes, government organizations and industries. All versions of AI principles are with different considerations covering different perspectives and making different emphasis. None of them can be considered as complete and can cover the rest AI principle proposals. Here we introduce LAIP, an effort and platform for linking and analyzing different Artificial Intelligence Principles. We want to explicitly establish the common topics and links among AI Principles proposed by different organizations and investigate on their uniqueness. Based on these efforts, for the long-term future of AI, instead of directly adopting any of the AI principles, we argue for the necessity of incorporating various AI Principles into a comprehensive framework and focusing on how they can interact and complete each other.

## Artificial Intelligence Principles: Different School of Thoughts

AI ethics and social impacts have drawn serious attentions and lots of policy frameworks have been brought up by various organizations. We confine our study to different AI principles (including guidelines, codes, and initiatives) pertaining to the general governance of AI. Typically, such principles are literally and explicitly documented in an item-by-item style, announced as an efforts to express the proposers' values and attitudes towards the understanding, development, and utilization of AI. Technically detailed discussions, including techniques oriented standards, are not included in this study. Traditional principles on robotics are also not included in this study.

Based on these considerations, we collected 27 proposals of AI principles to date. For each of the collected principles, we extract the texts of direct relevance to the author's points (in most cases this means the title words of the principles). We also include the necessary comments from the raw text to make the extracted text self-explanatory. Principle proposals are grouped by their backgrounds:

• **Principles from Academia, Non-profits and Non-Governmental Organizations**: (1) Asilomar AI Principles (FLI 2017). (2) General Principles in *Ethically Aligned Design, version 2*, by IEEE (IEEE 2017). (3) Principles for Algorithmic Transparency and Accountability by ACM (USACM 2017). (4) The Japanese Society for Artificial Intelligence Ethical Guidelines (JSAI 2017). (5) The Montreal Declaration for a Responsible Development of Artificial Intelligence (Montreal 2017). (6) Three ideas from the Stanford Human-Centered AI Initiative (HAI) (Stanford 2018). (7) Three Rules for Artificial Intelligence Systems by the CEO of Allen Institute for Artificial Intelligence (Etzioni 2017). (8) Harmonious Artificial Intelligence Principles (HAIP 2018). (9) Universal Guidelines for Artificial Intelligence (The Public Voice 2018). (10) Principles for the Governance of AI (The Future Society 2017). (11) Tenets of Partnership on AI (PAI 2016). (12) Top 10 Principles For Ethical Artificial Intelligence (UNI Global Union 2017). (13) AI Policy Principles (ITI 2017).

• **Principles from Governments**: (14) AI R&D Principles (MIC 2017). (15) Draft AI Utilization Principles (MIC 2018). (16) AI Code (House of Lords 2018). (17) Ethical principles and democratic prerequisites, European Group on Ethics in Science and New Technologies (EGE 2018).

• **Principles from Industry**: (18) DeepMind Ethics & Society Principles (DeepMind 2017). (19) OpenAI[1] Charter (OpenAI 2018). (20) AI at Google: Our Principles (Google 2018). (21) Microsoft AI Principles (Microsoft 2018). (22) Microsoft CEO's 10 AI rules (Nadella 2016). (23) Principles for the Cognitive Era (IBM 2017). (24) Principles for Trust and Transparency (IBM 2018). (25) Developing AI for Business with Five Core Principles (Sage 2017). (26) SAP's Guiding Principles for Artificial Intelligence. (SAP 2018). (27) Sony Group AI Ethics Guidelines (Sony 2018).

---

* These authors contributed equally to this study.

[1] OpenAI identifies itself as "a non-profit AI research company".

## Semantically Linking Various AI Principles

We aim to link various AI principles from the perspectives that they considered in common. Common perspectives may not use exactly the same word term, and semantically equivalent and similar terms should be considered.

We first identified a set of manually chosen keywords as the core terms, which belong to 10 general topics. We use word2vec representation of the word to find keywords with similar meanings. Google word vector trained from news[1] is used. The similarity between the original keyword and the other words is calculated by the cosine similarity between the word vector of the original keyword and the other words. A list of candidate extended keywords ranked by similarity is generated. The first word on the list with obviously deviated semantic meaning from the original keyword is selected as the threshold point, and all words with lower similarity are abandoned. Some phrases with similar meanings are added to the expanded keyword list. For example, for the term "collaboration", the expanded list also includes collaborations, collaborative, collaboratively, collaborate, collaborates and collaborating. While for the term "fairness", the expanded list also includes fair, fairer, unfair and unfairness.

*Table 1.Topics and Manually Chosen Keywords for AI Principles*

| Topics | Keywords |
|---|---|
| Humanity | humanity, beneficial, well-being, human value, human right, dignity, freedom, education, common good, human-centered, human-friendly |
| Collaboration | collaboration, partnership, cooperation, dialogue |
| Share | share, equal, equity, inequity, inequality |
| Fairness | fairness, justice, bias, discrimination, prejudice |
| Transparency | transparency, explainable, predictable, intelligible, audit, trace, opaque |
| Privacy | privacy, personal information, data protection, informed, explicit confirmation, control the data, notice and consent |
| Security | security, cybersecurity, cyberattack, hacks, confidential |
| Safety | safety, validation, verification, test, controllability, under control, control the risks, human control |
| Accountability | accountability, responsibility |
| AGI/ASI | AGI, superintelligence, super intelligence |

Here we define the topic coverage of a principle proposal as the percentage of topics that have been mentioned in the proposal. If any term or expanded keyword term has ever appeared in a proposal, we would mark that this proposal has covered the related topic. Table 1 presents 10 general topics and related terms for AI Principles. Term expansion efforts based on semantic similarities are introduced to extend the list for more comprehensive coverage.

---

[1] https://drive.google.com/file/d/0B7XkCwpI5KDYNlNUT-TlSS21pQmM/edit

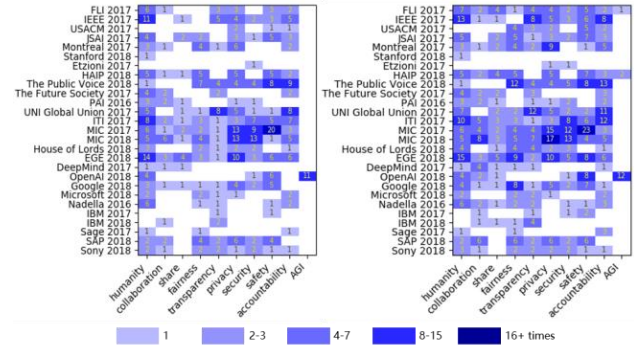

*Figure 1. Topic coverage of principle titles and explanatory texts based on manually chosen keywords (A) and extended keyword groups by semantic similarity (B).*

Figure 1 shows the coverage of different principles on the 10 topics. The colors are related to how many times the term appeared in the proposal. As can be observed, expanding the keywords using semantic similarity significantly increased topics found in principles, making the semantic analysis more accurate and robust against different use of similar word terms and expressions. The linkages among different AI principles are represented using Semantic Web standards (RDF/OWL) on the LAIP platform.

## Complementary Considerations from Different Organizations and Different AI Principles

Different principle proposals are compared by calculating their coverage on topics and keywords, as shown in Figure 2. We can observe that one of the principle proposals covered all the major topics. Among the top 10 proposals ranked by keywords, 8 of them ranked top 10 on topic coverage ranking as well. However, SAP 2018 ranked higher on keywords coverage ranking (the 10th), but ranked comparatively lower on topic coverage ranking (the 14th in parallel), since it discussed extensively about collaboration, fairness, privacy, and safety, while may have missed the topics of share, accountability and AGI/ASI. HAIP 2018 covered 8 of the 10 major topics (the 7th in parallel) without going through much of the details, hence ranked lower in keywords ranking (the 16th). We should emphasize that coverage of a proposal may not reflect lacking of considerations on certain topics, but just reflects that they may choose to have different emphasis. On the other hand, different considerations may interact to complement with each other.

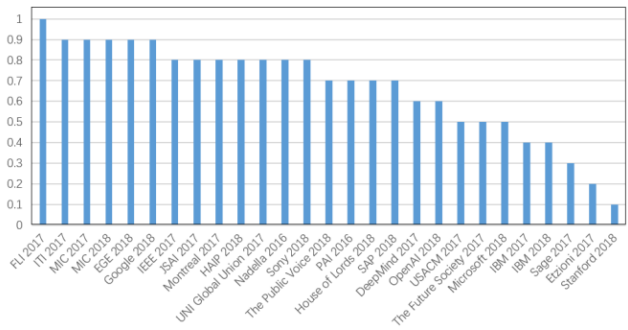

*(A) Topic Coverage Ranking*

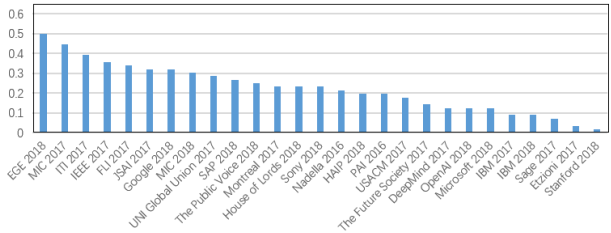

*(B) Keywords Coverage Ranking*

*Figure 2. Coverage Ranking for General Topics (A) and Keywords (B) for Different AI Principle Proposals*

According to the division of different school of thoughts from the types of publisher point of view, Figure 3 shows the comparative frequency of topics mentioned in three different types of AI principle proposals.

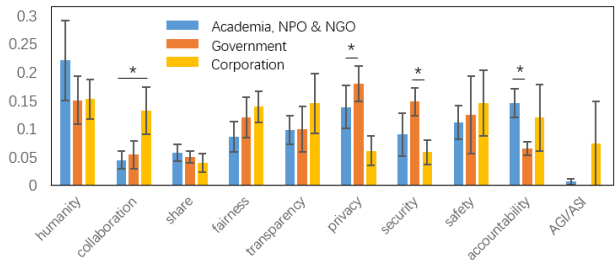

*Figure 3. Average topic frequency in different types of publishers, with standard error of the data. The asterisks indicate that the T-test p-value of the data is less than 0.05.*

We can observe from Figure 3 that corporations would like to mention more about collaboration, but not that much for security and privacy. While governments mentioned more about security, but would not like to mention accountability. Corporations can benefit from collaboration, but the atmosphere of collaboration may not be as good as academia, which may be the reason why they would like to mention it. Privacy and security are sensitive issues for corporations, maybe that is why corporations would not like to mention them. And the government mentioned the topic of accountability significantly less than academia.

Although in most cases, principles from different organizations usually share a common vocabulary, ambiguities in the analysis of the text still remain. The ambiguities may come from the polysemy of words and the context. For example, "race" is used in the context of "arms race" and "race avoiding" (FLI 2017) to represent the competition across researchers and nations (thus referring to the topic of "collaboration"), it is also used in the context of "gender, race, sexual orientation" (UNI Global Union 2017) to talk about possible biases of AI system (thus referring to the topic of "fairness"). Meanwhile, the "self-improvement" of an advanced AI system is a trait we should be very cautious about (FLI 2017), yet such "self-improvement" of AI researchers is what we ask for (JSAI 2017).

Such ambiguities also appear within a topic. For instance, we may ask for "transparency" from the decision-making process of the system out of our fairness concerns. We may also ask for "transparency" from the system to make it more safe, traceable, and controllable. The Asilomar AI principles have made such distinctions explicitly in their discussions (see "Judicial Transparency" and "Failure Transparency" in (FLI 2017)) while others usually seem to take one side of the concept or mixed them up. The ambiguities in these cases can be derived from the high-level abstraction of the concept itself and is also a reflection of the inner linkage between various topics.

Besides the general topics those AI Principle proposals share in common, many principles also reflect the unique perspectives of different organizations. For example, the Montreal Declaration has suggested promoting the well-being of "all sentient creatures", which according to their definition, includes "any being able to feel pleasure, pain, emotions; basically, to feel" (Montreal 2017). The JSAI Ethical Guidelines include that AI must abide these guidelines "in the same manner as the members of the JSAI in order to become a member or a quasi-member of society" (JSAI 2017). The General Principles from IEEE's report recommend that "For the foreseeable future, A/IS should not be granted rights and privileges equal to human rights: A/IS should always be subordinate to human judgment and control" (IEEE 2017). IBM takes the view that "Cognitive systems will not realistically attain consciousness or independent agency" and thus lay their stress on promoting AI and cognitive systems to "augment human intelligence" (IBM 2017). Those different perspectives from different proposals reflect the diversity of the whole AI community and it turns to be necessary to identify and incorporate such various considerations for a more comprehensive framework.

Based on the analysis, we have the following suggestions for future research and proposals for AI Principles:

- Strengthening safety-related considerations in academia and industry. Safety issues are the core for AI governance and have been realized in different government organizations, but many of the AI companies have not taken this seriously. While their AI products will directly bring potential risks for society.

- Long-term strategic design for AGI and ASI. Most AI principles investigated here do not cover considerations for AGI and ASI. While most of them should have been regarded as relatively long-term design for AI. Long-term planning on AGI and ASI will have clearer observations for strategic future and could have arrangements for potential risks in advance.
- From Human-centered to Harmonious Principle Design. Current AI principle proposals mainly focus on beneficial, human-centered design, while lack of considerations that the human society is on the way for transformation. More harmonious design considering both human and future AI as cognitive living systems should be considered.

## Conclusions

Different AI Principles have their own perspectives and coverage for the current and future strategies of AI. Instead of directly adopting any of the AI principles, we argue the necessity of linking and incorporating various AI Principles into a comprehensive framework and focusing on how they can interact and complement each other. The Linking Artificial Intelligence Principles (LAIP) platform is available as an online service under the address http://www.linking-ai-principles.org. It supports semantic search by keyword terms and paragraph search where semantically similar principles could be listed for exploration.

## Acknowledgement

This study is supported by New Generation of Artificial Intelligence Development Research Center, Ministry of Science and Technology of China under the project "Key issues of Social Ethics for Artificial Intelligence" from ISTIC.

## References

DeepMind. 2017. DeepMind Ethics & Society Principles. https://deepmind.com/applied/deepmind-ethics-society/principles/.

Etzioni, O. 2017. How to Regulate Artificial Intelligence. https://www.nytimes.com/2017/09/01/opinion/artificial-intelligence-regulations-rules.html.

European Group on Ethics in Science and New Technologies (EGE). 2018. Statement on Artificial Intelligence, Robotics and 'Autonomous' Systems. http://ec.europa.eu/research/ege/pdf/ege_ai_statement_2018.pdf.

Future of Life Institute (FLI). 2017. Asilomar AI Principles. https://futureoflife.org/ai-principles/.

Google. 2018. AI at Google: Our Principles. https://ai.google/principles.

HAIP Initiative. 2018. Harmonious Artificial Intelligence Principles (HAIP). http://bii.ia.ac.cn/hai/index.php.

House of Lords, UK. 2018. AI in the UK: ready, willing and able? https://publications.parliaent.uk/pa/ld201719/ldselect/ldai/100/100.pdf.

IBM. 2017. Principles for the Cognitive Era. https://www.ibm.com/blogs/think/2017/01/ibm-cognitive-principles/.

IBM. 2018. Principles for Trust and Transparency. https://www.ibm.com/blogs/policy/trust-principles/.

Information Technology Industry Council (ITI). 2017. AI Policy Principles. https://www.itic.org/public-policy/ITIAIPolicyPrinciplesFINAL.pdf.

Microsoft. 2018. Microsoft AI Principles. https://www.microsoft.com/en-us/ai/our-approach-to-ai.

Ministry of Internal Affairs and Communications (MIC), the Government of Japan. 2017. AI R&D Principles. http://www.soumu.go.jp//000507517.pdf.

Ministry of Internal Affairs and Communications (MIC), the Government of Japan. 2018. Draft AI Utilization Principles. http://www.soumu.go.jp//000581310.pdf.

Nadella, S. 2016. The Partnership of the Future: Microsoft's CEO explores how humans and A.I. can work together to solve society's greatest challenges. https://slate.com/technology/2016/06/microsoft-ceo-satya-nadella-humans-and-a-i-can-work-together-to-solve-societys-challenges.html.

OpenAI. 2018. OpenAI Charter. https://blog.openai.com/openai-charter/.

Partnership on AI (PAI). 2016. Tenets. https://www.partnershiponai.org/tenets/.

Sage. 2017. The Ethics of Code: Developing AI for Business with Five Core Principles. https://www.sage.com/ca/our-news/press-releases/2017/06/designing-AI-for-business.

SAP. 2018. SAP's Guiding Principles for Artificial Intelligence. https://news.sap.com/2018/09/sap-guiding-principles-for-artificial-intelligence/.

Sony. 2018. Sony Group AI Ethics Guidelines. https://www.sony.net/SonyInfo/csr_report/humanrights/hkrfmg0000007rtj-att/AI_Engagement_within_Sony_Group.pdf.

Stanford University. 2018. The Stanford Human-Centered AI Initiative (HAI). http://hai.stanford.edu/news/introducing_stanfords_human_centered_ai_initiative/.

The Future Society. 2017. Principles for the Governance of AI. http://www.thefuturesociety.org/science-law-society-sls-initiative/#1516790384127-3ea0ef44-2aae.

The IEEE Global Initiative on Ethics of Autonomous and Intelligent Systems. 2017. Ethically Aligned Design, Version 2. http://standards.ieee.org/develop/indconn/ec/autonomous_systems.html.

The Japanese Society for Artificial Intelligence (JSAI). 2017. The Japanese Society for Artificial Intelligence Ethical Guidelines. http://ai-elsi.org/wp-content/uploads/2017/05/JSAI-Ethical-Guidelines-1.pdf.

The Public Voice. 2018. Universal Guidelines for Artificial Intelligence. https://thepublicvoice.org/ai-universal-guidelines/.

UNI Global Union. 2017. Top 10 Principles For Ethical Artificial Intelligence. http://www.thefutureworldofwork.org/media/35420/uni_ethical_ai.pdf.

University of Montreal. 2017. The Montreal Declaration for a Responsible Development of Artificial Intelligence. https://www.montrealdeclaration-responsibleai.com/the-declaration.

US Public Policy Council, Association for Computing Machinery (USACM). 2017. Principles for Algorithmic Transparency and Accountability. https://www.acm.org/binaries/content/assets/public-policy/2017_usacm_statement_algorithms.pdf.